\pdfoutput=1
\documentclass[11pt,a4paper]{article}
\usepackage[hyperref]{emnlp2018}
\usepackage{times}
\usepackage{latexsym}
\usepackage{times}
\usepackage{latexsym}
\usepackage{times}
\usepackage{latexsym}
\usepackage{graphicx}
\usepackage{url}
\usepackage{subfigure}
\usepackage{tikz}
\usepackage{CJKutf8}
\usepackage{amsmath}
\usepackage{url}
\usepackage{url}

\aclfinalcopy 

\title{BERTSel: Answer Selection with Pre-trained Models}

\author{Dongfang Li \quad Yifei Yu \quad Qingcai Chen \quad Xinyu Li\\ Harbin Institute of Technology (Shenzhen) \\ {\tt crazyofapple@gmail.com} \\ {\tt yuyifei@stu.hit.edu.cn} \\ {\tt qingcai.chen@hit.edu.cn} \\ {\tt xinyulee.sz@gmail.com}}

\date{}

\begin{document}
\maketitle

\begin{abstract}

Recently, pre-trained models have been the dominant paradigm in natural language processing. They achieved remarkable state-of-the-art performance across a wide range of related tasks, such as textual entailment, natural language inference, question answering, etc. BERT, proposed by Devlin et.al., has achieved a better marked result in GLUE leaderboard with a deep transformer architecture. Despite its soaring popularity, however, BERT has not yet been applied to answer selection. This task is different from others with a few nuances: first, modeling the relevance and correctness of candidates matters compared to semantic relatedness and syntactic structure; second, the length of an answer may be different from other candidates and questions. In this paper. we are the first to explore the performance of fine-tuning BERT for answer selection. We achieved SOTA results across five popular datasets, demonstrating the success of pre-trained models in this task.
\end{abstract}

\section{Introduction}
Answer selection is the task of finding which of the candidates can answer the given question correctly. Since the emergency and development of deep learning methods in this task, the impressive results are yielded without relying on feature engineering or external knowledge bases~\cite{trecsota,trecqa1,wikiqa1,yahooqa1,semeval1,semeval2}.
However, many of them are based on shallow pre-trained word embedding and task-specific network structures. Until recently, the pre-trained language representation models, such as ELMo, GPT, and BERT~\cite{elmo2018,gpt,bert}. They achieve state of the art performance in many natural language processing tasks. 
In general, BERT is firstly pre-trained on vast amounts of text with expensive computational resources, by using two tasks: an unsupervised objective of masked language modeling and next-sentence prediction. Then, the pre-trained network is fine-tuned on task-specific labeled data.  However, BERT has not yet been fine-tuned for answer selection. In this paper, we explore fine-tuning BERT for this task. 
\begin{figure}
    \centering
    \includegraphics[width=0.4\textwidth,angle=0]{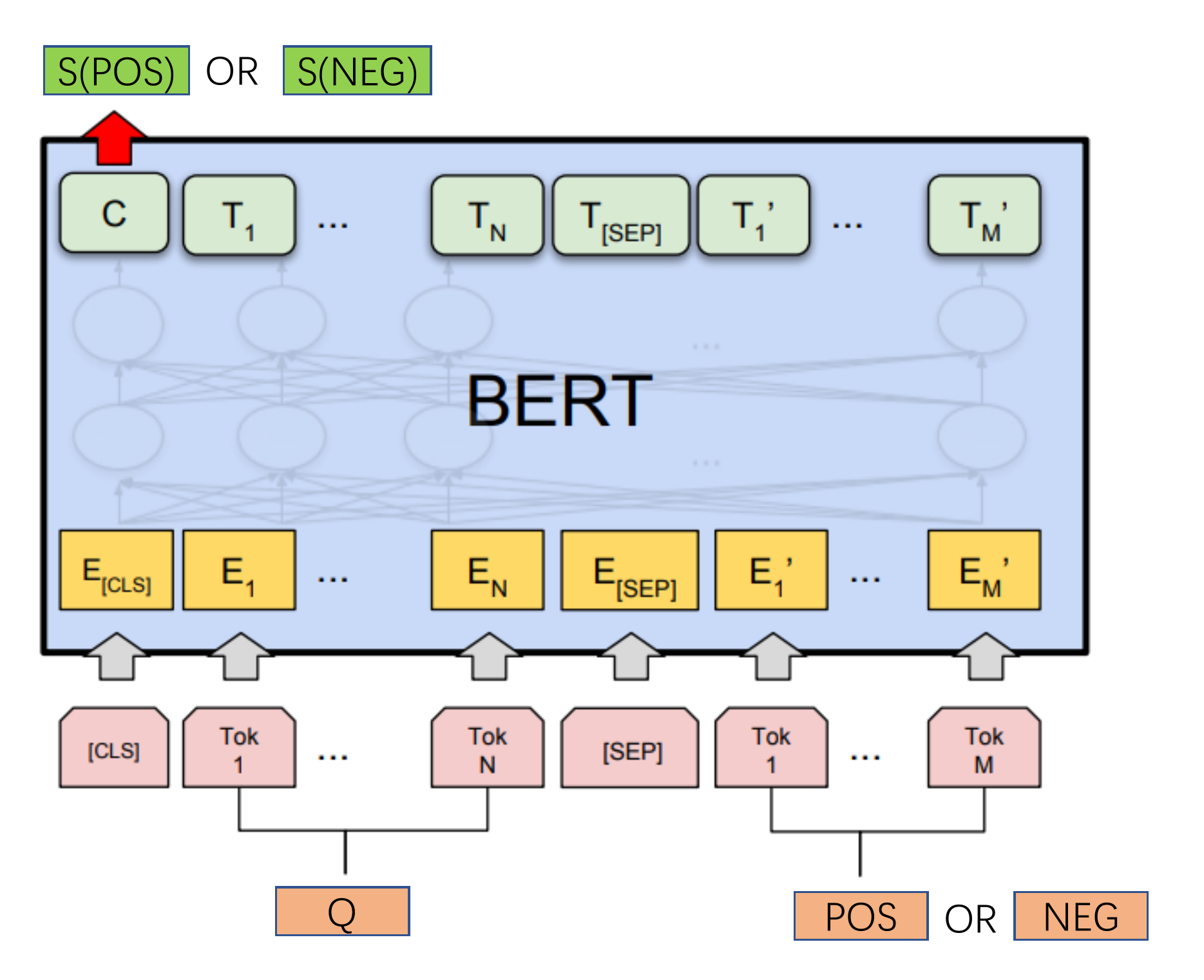}
    \caption{The fine-tuning method we used for answer selection task.}
    \label{fig:pipeline}
\end{figure}
Briefly, our contributions are: 
\begin{itemize}
	\item We explore the BERT pre-trained model to address the poor generalization capability of answer selection.
	\item We conduct exclusive experiments to analysis the effectiveness of BERT in this task, and achieve the state-of-the-art results in 5 benchmarks datasets.

\end{itemize}
\begin{table} 
  \centering
  \small
  
  \begin{tabular}{{c}{c}{c}{c}{c}}
    \hline
	Dataset & \#Train & \#Dev & \#Test & \#TrainPairs \\ \hline
	TrecQA & 1229 & 65 & 68 & 504421 \\
	WikiQA & 873 & 126 & 243 & 8995 \\
	YahooQA & 50112 & 6289 & 6283 & 253440 \\
	SemEvalcQA-16 & 4879 & 244 & 327 & 75181 \\
	SemEvalcQA-17 & 4879 & 244 & 293 & 75181 \\ \hline
  \end{tabular}
  \caption{Statistics of datasets of Answer Selection.}
  \label{table:data_stat}
\end{table}
\begin{table*}
    \centering
    \begin{tabular}{{c}{c}{c}{c}{c}{c}{c}{c}{c}{c}{c}{c}}
    \hline
    	 & \multicolumn{2}{c}{TrecQA} &  \multicolumn{2}{c}{WikiQA} & \multicolumn{2}{c}{YahooQA} & \multicolumn{2}{c}{SemEvalcQA-16} & \multicolumn{2}{c}{SemEvalcQA-17}  &  \\ \hline
    	 & MRR & MAP & MRR & MAP & MRR & MAP & MRR & MAP & MRR & MAP \\ \hline
    	epoch=3 & 0.927 & 0.877 & 0.770 & 0.753 & 0.942 & 0.942 & 0.872 & 0.810 & 0.951 & 0.909 \\ \hline
    	epoch=5 & 0.944 & 0.883 & 0.784 & 0.769 & 0.942 & 0.942 & 0.890 & 0.816 & 0.953 & 0.908 \\ \hline
    	SOTA & 0.865 & 0.904 & 0.758 & 0.746 & - & 0.801 & 0.872 & 0.801 & 0.926 & 0.887 \\ \hline
   \end{tabular}
   \caption{Results of BERT$_{base}$ in test set of five datasets with different epochs. The SOTA results are from~\cite{trecsota} (TrecQA), ~\cite{wikiqa1} (WikiQA, SemEvalcQA-16), ~\cite{trecqa1} (YahooQA), ~\cite{semeval2017} (SemEvalcQA-17). }
   \label{table:bert_base1}
\end{table*} 
\begin{table*}
    \centering
    \begin{tabular}{ {c}{c}{c}{c}{c}{c}{c}{c}{c}{c}{c}}
    \hline
    	 & \multicolumn{2}{c}{TrecQA} &  \multicolumn{2}{c}{WikiQA} & \multicolumn{2}{c}{YahooQA} & \multicolumn{2}{c}{SemEvalcQA-16} & \multicolumn{2}{c}{SemEvalcQA-17}\\ \hline
    	 & base & large & base & large & base & large & base & large & base & large \\ \hline
    	MRR & 0.927 &  \textbf{0.961} & 0.770 &  \textbf{0.875} &  \textbf{0.942} & 0.938 & 0.872 &  \textbf{0.911} & 0.951 &  \textbf{0.958} \\ \hline
    	MAP & 0.877 &  \textbf{0.904} & 0.753 &  \textbf{0.860} &  \textbf{0.942} & 0.938 & 0.810 &  \textbf{0.844} &  \textbf{0.909} & 0.907 \\ \hline
    \end{tabular}
    \caption{Results of BERT$_{base}$ and BERT$_{large}$ in test set of five datasets. The number of training epochs is 3.}
    \label{table:bert_large}
\end{table*} 
\section{Method}

Our method is based on the pairwise approach, as depicted on Figure~\ref{fig:pipeline}, which is taking a pair of candidate answer sentences and explicitly learns to predict which sentence is more relevant to the question. Given a question $q$, a positive answer $p$ and a sampled negative answer $q$, the model inputs are triples $(q, p, n)$. For fine-tuning, we split the triple into $(q, p)$ and $(q, n)$, and send them to BERT to get [CLS] embedding respectively. Then a fully connected layer and a sigmoid function are applied in each output logits to get final score.

\section{Experiments}

\subsection{Datasets}

TrecQA~\cite{trecqadataset}, WikiQA~\cite{wikiqadataset}, YahooQA~\cite{yahooqadataset} and SemEval cQA task~\cite{semeval2016, semeval2017} have been widely used for benchmarking answer selection models. Table~\ref{table:data_stat} summarizes the statistics of the datasets.
\subsection{Loss Function}
The training objective of the our model consists of two
aspects. For the first part, the objective function is to maximize the negative cross-entropy of positive and negative examples. For the second part, the objective function is hinge loss function.  By adding them, our final loss function
is obtained as follows:
\begin{equation}\label{loss function}
\begin{split}
\lambda_1(log\hat{y}_\theta(q,p) + log(1 - \hat{y}_\theta(q,n)) + \\ \lambda_2 max\{0, m - \hat{y}_\theta(q,p) + \hat{y}_\theta(q,n)\}
\end{split}
\end{equation}
where  $\hat{y}_\theta(q,p)$,  $\hat{y}_\theta(q,n)$ denote the predicted scores of positive and negative answer, $\lambda_{1} = 0.5$, $\lambda_{2} = 0.5$ are weighted parameters.

\subsection{Performance Measure}
The performance of an answer selection system is measured in Mean Reciprocal Rank (MRR) and Mean Average Precision (MAP), which are standard metrics in Information Retrieval and Question Answering. Given a set of question Q, MRR is calculated as follows:
\begin{equation}\label{mrr}
    MRR = \frac{1}{Q}\sum_{i=1}^{|Q|}\frac{1}{rank_i}
\end{equation}
where $rank_{i}$ refers to the rank position of the first correct candidate answer for the $i^{th}$ question. In other words, MRR is the average of the reciprocal ranks of results for the questions in Q. On the other hand, if the set of correct candidate answers for a question $q_j \in Q$ is $\{ d_1, d_2, ..., d_{m_j} \}$ and $R_{jk}$ is the set of ranked retrieval results from the top result until you get to the answer $d_k$, then MAP is calculated as follows:
\begin{equation}\label{map}
    MAP = \frac{1}{Q}\sum_{j=1}^{|Q|}\frac{1}{m_j}\sum_{k=1}^{|m_j|}Precision(R_{jk})
\end{equation}
When a relevant answer is not retrieved at all for a question, the precision value for that question in the above equation is taken to be 0.
Whereas MRR measures the rank of any correct answer, MAP examines the ranks of all the correct answers.
\subsection{Results}
We show the main result in Table~\ref{table:bert_base1} and~\ref{table:bert_large}. Despite training on a fraction of the data available, the proposed BERT-based models surpass the previous state-of-the-art models by a large margin on all datasets.

\section{Conclusion}

Answer selection is an important problem in natural language processing, and many deep learning methods have been proposed for the task. In this paper, We have described a simple adaptation of BERT that has become the state of the art on five datasets.

\bibliography{emnlp2018}
\bibliographystyle{acl_natbib_nourl}

\end{document}